\pdfoutput=1

\documentclass[10pt,twocolumn,letterpaper]{article}

\usepackage{cvpr}      %

\usepackage{float}

\definecolor{cvprblue}{rgb}{0.21,0.49,0.74}
\usepackage[pagebackref,breaklinks,colorlinks,allcolors=cvprblue]{hyperref}

\def\paperID{*****} %
\def\confName{CVPR}
\def\confYear{2026}
 
\title{AUTOPILOT VQA: Benchmarking Vision-Language Models for Incident-Centric Dashcam Understanding}
\author{
Siddharth Damodharan\\
University of Colorado Colorado Springs
\and
Radhika Gupta\\
University of Michigan
\and
Ali Alshami\\
University of Colorado Colorado Springs
\and
Ryan Rabinowitz\\
University of Notre Dame
\and
Jugal Kalita\\
University of Colorado Colorado Springs
}

\begin{document}
\maketitle
\begin{abstract}
Recent advances in Vision-Language Models, Large Language Models, and Multimodal Large Language Models have improved autonomous driving tasks such as scene understanding, decision making, trajectory prediction, and visual question answering. However, evaluating whether these models can reliably reason about safety-critical incidents remains challenging. To address this gap, we present AUTOPILOT-VQA, an incident-centric visual question answering benchmark for dashcam video understanding. The dataset evaluates different systems through structured questions designed around real-world driving incidents and near-incidents. The benchmark covers diverse safety-relevant categories, including weather and lighting conditions, traffic environment, road layout, road surface state, signage, involved entities, accident occurrence, impact location, and avoidability-related reasoning. By requiring models to answer grounded questions about both contextual scene properties and event-level incident details, AUTOPILOT-VQA moves beyond object recognition toward temporally grounded, safety-aware reasoning. The dataset is released as part of the AUTOPILOT CVPR 2026 competition and provides a standardized benchmark for assessing the reliability of autonomous driving systems in different scenarios. Our benchmark support developments for more interpretable, robust, and safety-conscious vision-language systems for real-world autonomous driving.

\end{abstract}
    
\section{Introduction}
\label{sec:intro}

\begin{figure}[t]
\centering
\includegraphics[width=0.50\textwidth]{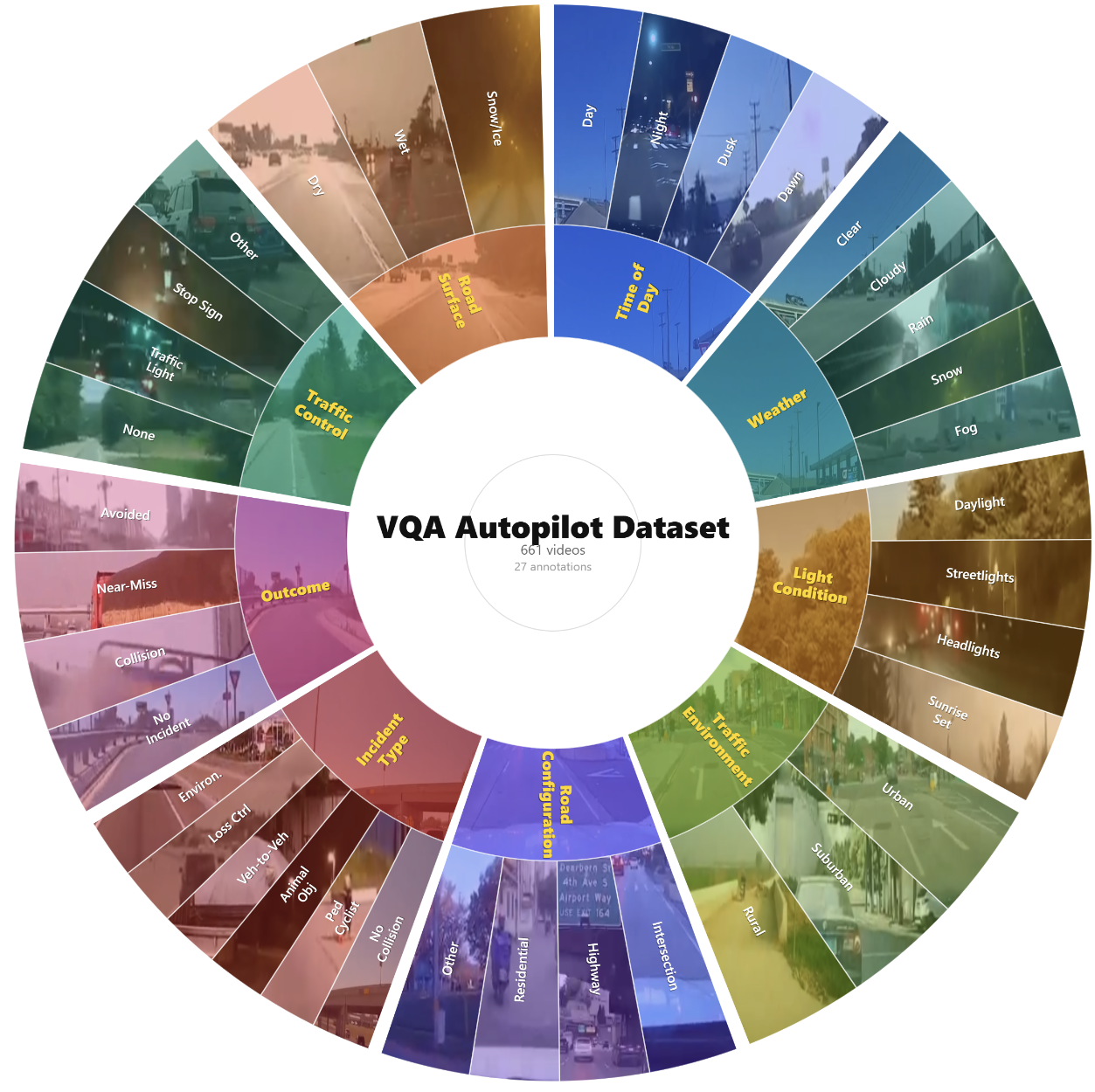}
\caption{
Overview of the VQA-Autopilot dataset annotation schema. The visualization illustrates the hierarchical structure of annotation categories, including environmental conditions, traffic context, incident types, outcomes, and associated attributes. This figure provides a high-level conceptual summary of the dataset design.
}
\label{fig:dataset_overview}
\end{figure}

Autonomous vehicles rely on accurate interpretation of complex traffic scenes to ensure safe and reliable operation in real-world environments. In everyday driving, autonomous systems must process large volumes of multimodal sensor data while responding to dynamic and uncertain conditions, including surrounding traffic behavior, road geometry, pedestrians, obstacles, and weather or visibility changes. Errors in scene understanding can propagate to planning and control, leading to unsafe or incorrect driving decisions. Consequently, robust perception and contextual reasoning remain central challenges in autonomous vehicle development~\cite{geiger2012kitti,janai2017computer,caesar2020nuscenes}.

Substantial progress has been made in autonomous driving over the past decade, largely driven by large-scale public datasets and standardized benchmarks. Open-source resources such as KITTI~\cite{geiger2012kitti}, Cityscapes~\cite{cordts2016cityscapes}, BDD100K~\cite{yu2020bdd100k}, nuScenes~\cite{caesar2020nuscenes}, Argoverse~\cite{chang2019argoverse}, and the Waymo Open Dataset~\cite{sun2020waymo} have accelerated research in object detection, tracking, lane understanding, localization, and trajectory forecasting. These benchmarks have enabled modern autonomous systems to achieve strong performance under routine driving conditions and structured urban environments.

Despite these advances, rare safety-critical incidents remain among the most challenging scenarios for autonomous driving systems. Accidents, near-misses, and other hazardous events often arise from multiple interacting factors rather than a single perception failure. Understanding such scenes requires more than detecting visible objects or classifying an event from video alone. A system must reason about the surrounding context, identify the entities involved, interpret road-user behavior, assess severity, and infer factors relevant to emergency response or preventive action. In addition, accident understanding often benefits from language-based information, such as natural-language scene descriptions or police-report-style summaries, which can provide details about fault, severity, involved parties, and incident progression. Because these events are infrequent but disproportionately consequential, structured accident-scene understanding remains an important open challenge in autonomous driving~\cite{ramanishka2018toward,fang2023vision}.

To address this challenge, we introduce \textit{Autopilot VQA}, a benchmark for structured accident-scene understanding in autonomous driving. The dataset contains more than 600 dashcam video clips spanning collisions, near-misses, and no-incident baseline sequences. Each clip is annotated through six semantic groups, resulting in more than 6,000 question-answer pairs that cover environmental conditions, road configuration, involved-entity behavior, incident category, fault attribution, and impact characterization. The benchmark is designed to evaluate whether models can reason beyond low-level perception and produce structured answers about safety-critical driving scenarios.

We further establish a public evaluation framework through a Kaggle competition, enabling participants to develop and compare diverse modeling pipelines under a shared scoring protocol. This setting supports reproducible evaluation while encouraging progress on multimodal reasoning for accident analysis.

Overall, our contributions are threefold. First, we introduce an accident-centric visual question answering benchmark with structured annotations for safety-critical traffic scenes. Second, we provide an open and competitive evaluation platform for reproducible comparison across methods. Third, we present empirical results that highlight the limitations of current models in contextual and safety-critical reasoning for autonomous driving.

\section{Related Work}

Large-scale datasets and standardized benchmarks have been crucial to advancing computer vision for autonomous driving systems, particularly in perception, tracking, localization, and traffic scene understanding~\cite{janai2017computer}. However, many of these tasks focus only on routine driving conditions, while higher-level reasoning in complex, safety-critical environments remains unexplored. Furthermore, understanding agentic interactions, hazardous behaviors, and incident context requires capabilities beyond those provided by common perception frameworks.

Recent work has begun to explore the application of VLMs to drive specific contexts through structured evaluation frameworks. For example, DrivingVQA introduces a visual question-answering data set designed to evaluate the reasoning of the model in real-world driving scenarios, highlighting that current models struggle with multimodal reasoning and rely heavily on salient visual cues~\cite{corbiere2026drivingvqa}. Furthermore, MetaVQA proposes a benchmark for evaluating spatial reasoning and scene understanding in autonomous driving, demonstrating that advanced models exhibit limited spatial awareness and difficulty reasoning about dynamic environments~\cite{metavqa}.

More recent benchmarks, such as NuPlanQA, provide large-scale evaluation frameworks to drive scene understanding by decomposing tasks into perception, spatial reasoning, and decision-making components. These works highlight that current vision-language models struggle to generalize to complex driving environments, particularly for tasks requiring structured reasoning and contextual understanding~\cite{park2025nuplanqa}. 

While prior work primarily focuses on model development or static benchmark evaluation, our work develops a structured dataset focused on accident scenarios. It evaluates performance through a competitive, open benchmarking framework. By leveraging a Kaggle competition, we obtain diverse modeling approaches and gain broader insights into how different methods perform on safety-critical driving tasks.

\section{Dataset and Annotation}

\subsection{Source Material}

Our evaluation is built on the VQA-Autopilot dataset, which contains over 600 dashcam video clips capturing a wide range of driving scenarios, from uneventful highway driving to multi-vehicle collisions. Critically, the dataset covers the full incident severity spectrum: approximately 27\% of clips involve direct collisions, 11\% near-misses, 17\% situations where a hazard was avoided in advance, and 27\% no-incident baseline sequences (see Figure 1). This balanced coverage prevents models from achieving high accuracy through class prior exploitation alone.

\subsection{Annotation Schema}

Each video was labeled by human annotators across nine structured question groups (A--I), yielding 28 sub-questions in total. The categories span environmental conditions (weather, time of day, lighting), road context (traffic environment, road configuration, lane structure, road surface, traffic control presence), and incident characterization (incident category, primary and secondary entity identity and behavior, fault attribution, prevention measures, and impact location). In total, this produces over 6,000 annotated question-answer pairs across the dataset.

\subsection{Dataset Statistics}
\begin{figure*}[!t]
\centering
\includegraphics[width=\textwidth]{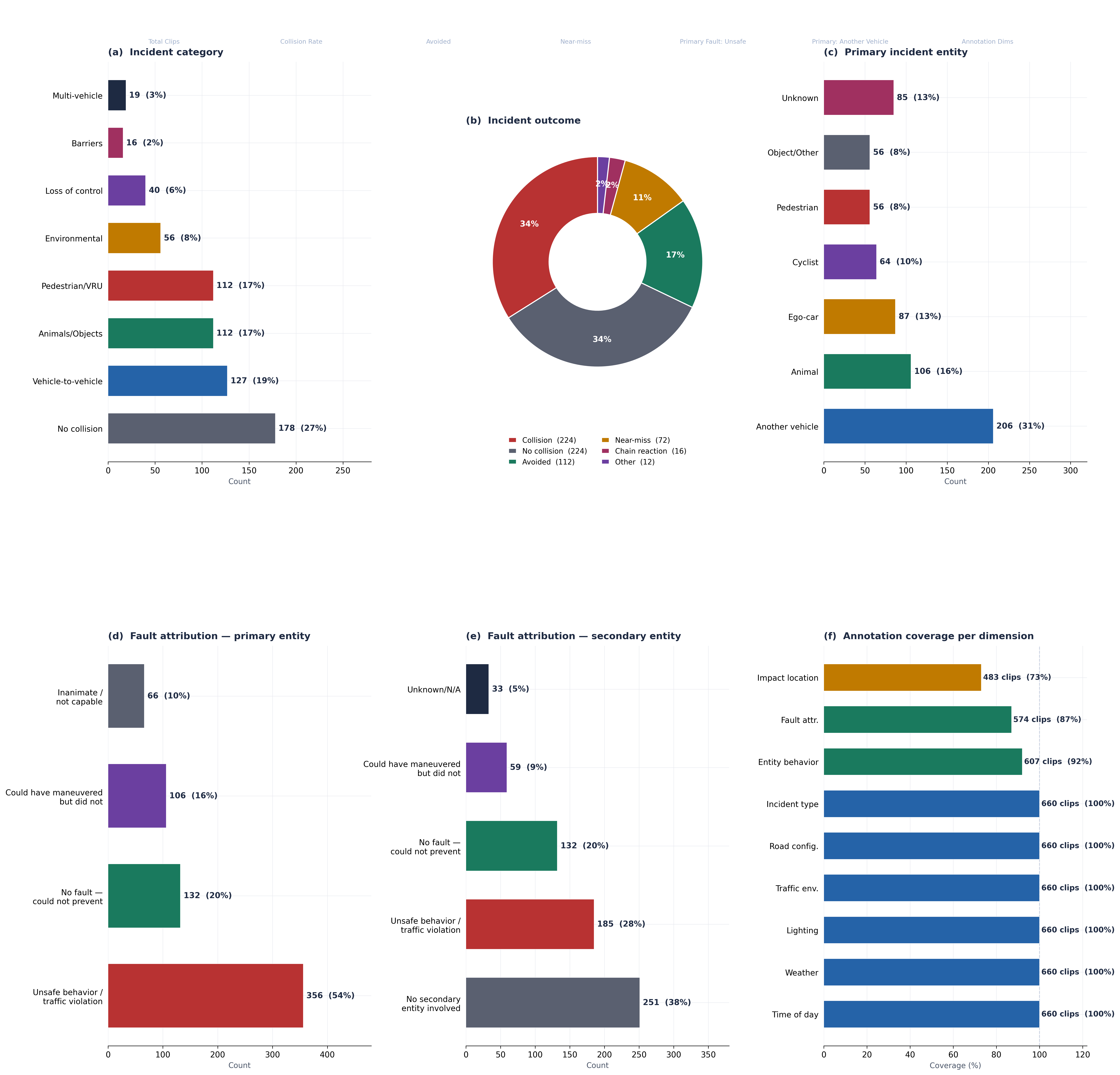}
\caption{
Environmental context statistics of the VQA-Autopilot dataset. This figure summarizes distributions across time of day, weather conditions, road surface state, road configuration, traffic control presence, and their co-occurrence. These distributions reflect realistic driving conditions and highlight the diversity required for evaluating contextual scene understanding.
}
\label{fig:env_context}
\end{figure*}

As shown in Figures A--F, the dataset captures realistic diversity across all annotation dimensions. The majority of incidents occur during daytime (70\%) under clear or partly cloudy conditions (68\%), reflecting naturalistic dashcam deployment. Highway and intersection scenarios together account for 65\% of the dataset. Primary incident entities include other vehicles (31\%), animals (16\%), ego-vehicle fault scenarios (13\%), and vulnerable road users, including cyclists and pedestrians (18\%). Road surfaces are predominantly dry (83\%), with wet conditions present in 17\% of clips, predominantly in rainy weather sequences.

\section{Kaggle Competition and Results}

\subsection{Task Definition}
The Hosted Kaggle Competition is formatted as a structured Visual Question Answering (VQA) task on dashcam video data. Within a given driving video, depicting both near-indecent and incident scenarios, participants are required to predict answers to a predefined set of questions that capture specific aspects of the scene. 

The questions participants are given span different dimensions within the driving context, including environmental conditions (weather and time), scene context (traffic environment and road configuration), and incident-specific attributes (e.g., involved parties, incident behaviors, and causes of incidents). In addition to the included tasks, there are higher-level reasoning tasks, such as identifying potential incident-prevention measures and specific impact points. 

For each video, participants must output a set of fixed, integer-valued predictions corresponding to a predefined answer category. This structure enables consistency across submissions and ensures fine-grained evaluation of model performance and understanding

Overall, the task is designed to evaluate both perceptual capabilities and higher-level reasoning in safety-centered driving scenarios, with an emphasis on accident understanding.

\subsection{Evaluation Metrics}

Submissions are evaluated using mean per-question accuracy across all visual question answering (VQA) fields. For each question, accuracy is calculated as the proportion of correct predictions across the test set.

Let $Q$ denote the total number of questions and let $\mathrm{Acc}q$ represent the accuracy for question $q$. The final score is computed as:
\[
\text{Score} = \frac{1}{Q} \sum_{q=1}^{Q} \mathrm{Acc}_q
\]

Each question is weighted equally in the final score, this allow for a balanced evaluation across different aspects of scene understanding. Predictions correspond to predefined answer categories, allowing for consistent comparison across various models and approaches.

\subsection{Leaderboard Overview}
The AUTOPILOT VQA challenge attracted strong community participation, with 224 registered entrants, 73 active participants, 59 teams, and 686 total submissions. This level of engagement indicates substantial interest in safety-centered multimodal reasoning tasks. 

At the close of the public leaderboard phase, the top-performing team achieved a score of 0.65835, with the next two teams scoring 0.65505 and 0.65371 respectively. The narrow margin separating the top submissions suggests that performance gains near the leaderboard ceiling were increasingly difficult to obtain and likely required careful optimization, ensembling, or stronger multimodal representations.

A broader review of the leaderboard reveals a long-tailed distribution of scores. While several teams exceeded 0.60 accuracy, many submissions clustered near 0.39--0.40, indicating that naive baselines or partially optimized systems struggled to generalize across the full question set.

\subsection{Error Analysis}
Despite competitive leaderboard performance, the moderate top score indicates that many question categories remain challenging. Errors are likely concentrated in tasks requiring causal inference or fine-grained relational reasoning rather than simple perception.

For example, identifying weather, time of day, or broad traffic environment may be solvable through strong visual recognition alone. In contrast, predicting which entity could have prevented the incident, determining fault-relevant behaviors, or estimating likely impact regions requires temporal understanding and reasoning over interactions between multiple agents.

The benchmark also includes unknown and non-applicable classes, which introduce additional complexity. Models must learn not only to classify visible evidence, but also to recognize when insufficient information is present or when a question does not apply to the scenario.

\subsection{Observed Trends}
Several trends emerge from the leaderboard results. First, there appears to be a clear separation between highly competitive systems and the remainder of the field, with only a limited number of teams surpassing 0.60 mean accuracy. This suggests that strong performance likely depends on advanced model adaptation rather than direct zero-shot prompting.

Second, the dense clustering among top teams implies diminishing returns near current performance limits. Small leaderboard improvements may correspond to significant engineering effort, indicating that the benchmark remains challenging even for strong methods.

Third, the wide score variance across teams suggests that model selection, preprocessing strategy, and handling of structured outputs play an important role. Since submissions require multiple synchronized predictions per video, robust output formatting and task decomposition may materially affect final results.

\subsection{Conclusion}
The AUTOPILOT VQA competition demonstrates both the promise and current limitations of multimodal models for autonomous driving understanding. While top systems achieved meaningful performance, no method approached near-human reliability, particularly given the safety-critical nature of the task.

These results suggest that current VLM pipelines remain stronger at perception than at structured reasoning. Future progress may require methods that explicitly model temporal causality, agent interactions, uncertainty calibration, and hierarchical decision-making.

Additionally, competitive benchmark settings such as Kaggle provide a valuable mechanism for rapidly evaluating diverse approaches from the broader research community. By combining open participation with safety-relevant tasks, such challenges can accelerate progress toward more robust autonomous driving intelligence.

{
    \small
    \bibliographystyle{ieeenat_fullname}
    \bibliography{main}
}

\end{document}